\documentclass[review]{elsarticle}%
\usepackage{algorithm}%
\usepackage{algorithmicx}%
\usepackage{algpseudocode}%
\usepackage{graphicx}%
\usepackage{subfig}%
\usepackage{xcolor}%
\usepackage{amssymb}%
\usepackage{amsmath}%
\usepackage{url}%
\begin{document}%
\begin{frontmatter}
\title{%
Skeleton Based Action Recognition using a Stacked Denoising Autoencoder with Constraints of Privileged Information
}%
\author[hfuu]{Zhi-ze Wu}%
\ead{wuzhize@mail.ustc.edu.cn}%
\ead[url]{iao.hfuu.edu.cn}%
\author[hfuu]{Thomas Weise}%
\ead{tweise@gmx.de}%
\ead[url]{iao.hfuu.edu.cn}%
\author[hfuu2]{Le Zou}%
\ead{zoule@hfuu.edu.cn}%
\author[hfuu2]{Fei Sun\corref{mycorrespondingauthor}}%
\cortext[mycorrespondingauthor]{Corresponding author}
\ead{sunfei@hfuu.edu.cn}%
%
\author[hfuu2]{Ming Tan}%
\ead{tanming@hfuu.edu.cn}%
\address[hfuu]{%
Institute of Applied Optimization, %
School of Artificial Intelligence and Big Data, %
Hefei University; %
Jinxiu Dadao 99, %
Hefei, Anhui, China, 230601.%
}%
\address[ustc]{%
School of Computer Science and Technology, %
University of Science and Technology of China; %
Hefei, Anhui, China, 230027.%
}%
\address[hfuu2]{%
School of Artificial Intelligence and Big Data, %
Hefei University; %
Jinxiu Dadao 99, %
Hefei, Anhui, China, 230601.%
}%
\begin{abstract}%
Recently, with the availability of cost-effective depth cameras coupled with real-time skeleton estimation, the interest in skeleton-based human action recognition is renewed.
Most of the existing skeletal representation approaches use either the joint location or the dynamics model. 
Differing from the previous studies, we propose a new method called Denoising Autoencoder with Temporal and Categorical Constraints \mbox{(DAE\_CTC)} to study the skeletal representation in a view of skeleton reconstruction.
Based on the concept of learning under privileged information, we integrate action categories and temporal coordinates into a stacked denoising autoencoder in the training phase, to preserve category and temporal feature, while learning the hidden representation from a skeleton. 
Thus, we are able to improve the discriminative validity of the hidden representation.
In order to mitigate the variation resulting from temporary misalignment, a new method of temporal registration, called Locally-Warped Sequence Registration (LWSR), is proposed for registering the sequences of inter- and intra-class actions. 
We finally represent the sequences using a Fourier Temporal Pyramid (FTP) representation and perform classification using a combination of LWSR registration, FTP representation, and a linear Support Vector Machine (SVM). 
The experimental results on three action data sets, namely MSR-Action3D, UTKinect-Action, and Florence3D-Action, show that our proposal performs better than many existing methods and comparably to the state of the art.%
\end{abstract}%
\begin{keyword}%
Action Recognition \sep Privileged Information \sep Denoising Autoencoder \sep Temporal Registration%
\end{keyword}%
\end{frontmatter}%
\section{Introduction}%
\label{section:introduciton}
Human action recognition (HAR) is a broad field of study concerned with identifying the movements or actions of a person in a given scenario automatically. 
HAR has a wide range of applications, including  surveillance systems, human-computer interaction, assistive technology, sign language recognition, computational behavior science, and consumer behavior analysis~\cite{wang2012mining,yang2012recognizing,wu2015discriminative}.
Essentially, it is a challenging time series classification task, which involves deep domain expertise and methods from signal processing to correctly engineer features from the raw data in order to fit a machine learning model~\cite{du2015hierarchical}.

The traditional HAR methods are mainly based on monocular RGB~videos recorded by 2D~cameras.
However, for the problems of variations in the view-point, illumination changes,  background clutter, and occlusions in human action analysis, the monocular RGB~data is difficult to provide effective support.
In addition, in terms of imaging sensors, monocular video sensors cannot provide 3D~space information about human movements~\cite{vemulapalli2014human,wu2018autoencoder}.
Therefore, despite much research invested over the past few decades, extrapolating advanced knowledge from RGB~videos, especially in complex and unconstrained scenarios, remains a challenging problem.

In 1975, Johansson carried out the famous experiment about moving light spots~\cite{johansson1975visual}, which opened up a new idea of human action recognition.
The experiment showed that the geometric structure of the human motion pattern is determined by the human skeleton.
Raising the reliability of the position and motion extraction of the human skeleton from video data can improve the effectiveness of action recognition~\cite{knutzen1998kinematics}.
With the development of depth imaging technology, skeletal data extraction algorithms based on depth images have become more mature and stable~\cite{shotton2013real,tompson2014joint}.
Especially, with the availability of cost-effective depth cameras coupled with real-time skeleton estimation like Kinect, the interest in skeleton-based human action recognition is rising.

Generally, most of existing skeleton-based human action recognition approaches~\cite{yang2012eigenjoints,wang2012mining,hussein2013human,eweiwi2014efficient,du2015hierarchical,luo2013group,wu2014leveraging} focus on the skeleton representation problem. 
For example, \cite{yang2012eigenjoints,wang2012mining,hussein2013human,eweiwi2014efficient} target joint-based features by simply taking the human skeleton as a set of points and attempting to model the motion of either individual joints or combinations of joints. 
Several works~\cite{du2015hierarchical,luo2013group,wu2014leveraging} model the dynamics of either subsets or all the joints in the skeleton using linear dynamical systems (LDSes)~\cite{slama2015accurate,chaudhry2013bio}, hidden Markov models (HMMs)~\cite{wu2014leveraging}, recurrent neural networks (RNNs)~\cite{du2015hierarchical}, or mixed approaches~\cite{presti2014gesture}.

Moreover, with the development of deep learning~\cite{lecun2015deep,krizhevsky2012imagenet,hochreiter1997long,vincent2010stacked}, the emergence of large-scale visual data sets~\cite{russakovsky2015imagenet,lin2014microsoft}, and the rapid update of hardware resources (such as GPU, FPGA, etc.), great progress has been achieved in traditional visual understanding tasks such as image classification~\cite{he2016deep}, object detection~\cite{ren2017faster,liu2016ssd}, and semantic segmentation~\cite{long2015fully,chen2017deeplab} and their accuracy now even exceeds that of humans.
Recently, there are various attempts on skeleton based action recognition using deep models~\cite{du2015hierarchical,zhu2016co,ke2017new,weng2017spatio}.
The most widely used models in deep-learning-based methods are convolutional neural networks (CNNs) and recurrent neural networks (RNNs), where the coordinates of joints are represented as vector pseudo-images or sequences, respectively.

Even if the positions of skeleton joints can be estimated~\cite{shotton2013real,tompson2014joint}, these estimates may be not reliable when the human body is only partly in view, the background is cluttered, or limbs overlap during the actions.
Unfortunately, the above factors, are common in practical applications.
The aforementioned approaches and models may fail when facing incompletely extracted positions of skeleton joints.
Therefore, more effective and efficient skeleton representation methods are required for real-word applications.
In order to allow for noisy estimates of the skeleton joint positions, we incorporate information of action categories and temporal coordinates into a skeleton representation algorithm from the perspective of skeleton reconstruction.
We therefore apply denoising autoencoders~\cite{vincent2010stacked,xing2016stacked} as an unsupervised feature-learning scheme in which the internal layer acts as a generic feature extractor of inner skeleton representations.

Rather than elaborately designing joint-based local features or dynamics models, we propose to learn discriminative representations from the skeleton data directly, including noisy and incomplete skeletons using a new deep architecture called Stacked Denoising Autoencoder with Temporal and Categorical Constraints \mbox{(SDAE\_CTC)}.
The learning under privileged information paradigm provides a model trained with additional information available only in the training phase and not at test time ~\cite{crasto2019mars}. 
In our case, action categories and temporal coordinates  are the privileged information available for training, along with skeleton, but only skeleton data is available at test time.
The proposed basic autoencoder architecture consists of three parts, namely skeleton reconstruction, action category reconstruction, and relative temporal reconstruction.
To obtain this model, we optimize the mapping parameters by applying a greedy layer-wise training process with labeled data.

\begin{figure}[tb]%
\centering%
\includegraphics[width=0.977\linewidth]{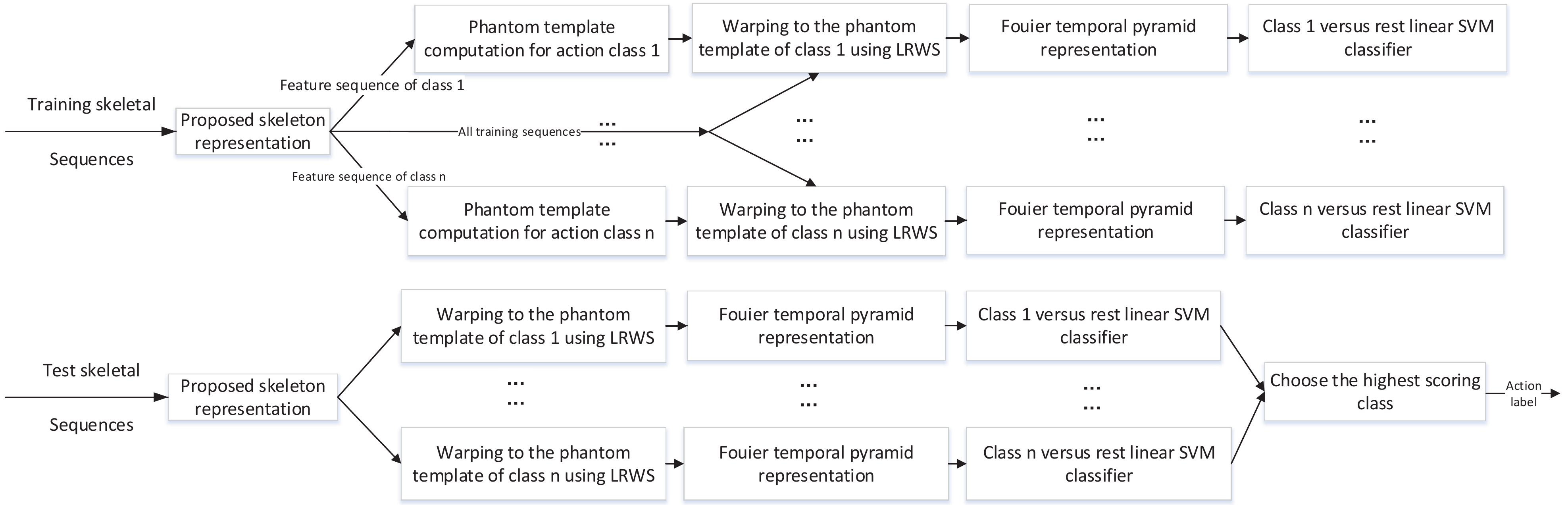}%
\caption{The top row shows all the steps involved in training and the bottom row shows all the steps involved in testing.}%
\label{fig:1}%
\end{figure}

Irrespective of the skeleton representation being used, there is an another fundamental yet challenging problem in human action recognition: How to deal with its temporal variations.
In order to mitigate temporary misalignments, we propose a new temporal registration method, called Locally-Warped Sequence Registration (LWSR).
The LWSR employs two functions named \mbox{$\mathit{LWSR\_intra}$} and \mbox{$\mathit{LWSR\_inter}$} to warp the intra-\ and inter-class sequences and generate a phantom action template of each action category.
Then, we represent the warped sequences using the Fourier temporal pyramid (FTP) representation proposed in~\cite{wang2012mining}.
Finally, we can perform the action classification by combining LWSR, FTP, and a linear Support Vector Machine (SVM).
Figure~\ref{fig:1} presents an overview of this proposed framework.%

\textbf{Contributions:} This paper makes four main contributions.
First, a new skeleton representation algorithm is proposed.
Second, a new method for temporal registration is designed to register temporal sequences.
Third, experimental results on three different data sets, including \mbox{MSR-Action3D~\cite{li2010action}}, \mbox{UTKinect-Action~\cite{xia2012view}}, and \mbox{Florence3D-Action~\cite{seidenari2013recognizing}}, show that our proposed method performs better than many existing skeletal representations and comparable to the state-of-the-art solutions, which are coupled with the much more intricate training architecture.
Finally, we experimentally show that our proposal is able to restore corrupted skeleton data.

\textbf{Organization:} The remainder of this paper is organized as follows.
We review the closely related work in Section~\ref{sec:2}, and specify  Denoising Autoencoders in Section~\ref{sec:3}. Section~\ref{sec:4}  formulates the proposed skeletal representation learning architecture. Section~\ref{sec:5} describes the temporal registration and classification approach. Experimental results and discussions are presented in Section~\ref{sec:6}.
Finally, we conclude the paper in Section~\ref{sec:7}.%
\section{Related Work}%
\label{sec:2}%
Action recognition based on skeleton data has attracted considerable attention, due to its effective representation of motion dynamics and strong adaptability to complicated background.
From the perspective of feature representation, we can divide existing algorithms into two approaches: the hand-crafted-based and the deep-learning-based.

The hand-crafted-based approaches~\cite{vemulapalli2016rolling,vemulapalli2014human,wang2012mining,hussein2013human} design algorithms to capture action patterns based on the physical intuitions.
Based on multiple kernel learning, Wang et al.~\cite{wang2012mining} propose to select discriminative joint combinations by mining actionlet ensemble. 
Vemulapalli et al.~\cite{vemulapalli2016rolling} represent each skeleton using the relative 3D rotations between various body parts.
The relative 3D geometry between all pairs of body parts is applied to represent the 3D human skeleton in~\cite{vemulapalli2014human}.
In order to encode the relationship between joint
movement and time, Hussein et al.~\cite{hussein2013human} deploy multiple covariance
matrices over sub-sequences in a hierarchical fashion.

With the development of deep learning, the deep-learning-based approaches have become the mainstream, where the most widely used models are LSTMs and CNNS. 
LSTM-based models~\cite{du2015hierarchical,weng2017spatio,zhu2016co} capture the temporal dependencies between consecutive frames.
In~\cite{du2015hierarchical}, skeleton joints are divided into five sets corresponding to five body parts.
They are fed into five LSTMs for feature fusion and classification.
In~\cite{zhu2016co}, the skeleton joints are fed to a deep LSTMs at each time slot to learn the inherent co-occurrence features of skeleton joints.
In~\cite{weng2017spatio}, both the spatial and temporal information of skeleton sequences are learned with a spatial-temporal LSTMs.
These methods first enrich the varieties of skeleton structure and then adopt LSTMs for action recognition.
CNN-based methods~\cite{ke2017new,si2018skeleton,li2018co,soo2017interpretable,liu2017enhanced} also achieve remarkable results.
Ke et al.~\cite{ke2017new} transform skeleton sequences to video clips and use a pre-trained CNN model to learn long-term temporal information of the skeleton sequence from the frames of the generated clips.
In~\cite{si2018skeleton}, Si et al. propose a novel model with spatial reasoning and temporal stack learning for skeleton-based action recognition.
Based on the capability to global aggregation of CNN, Li et al.~\cite{li2018co}  propose an end-to-end convolutional co-occurrence feature learning framework.
In~\cite{liu2017enhanced}, Liu et al. map skeleton joints into a view invariant high dimensional space and visualize them as color images. 
Based on such color images, a multi-stream CNN-based model is applied to extract and fuse deep features.
Soo et al.~\cite{soo2017interpretable} propose to use Temporal Convolutional Neural Networks to explicitly learn readily interpretable spatio-temporal representations for 3D human action recognition. 

The scenarios and prerequisites in these methods are different from our method.
LSTM-based approaches jointly consider the skeleton representation and temporal classification, which makes the whole recognition process much more intricate.
CNN-based approaches generally generate the pseudo-images first,  but this process will lose the 3D information of the action.
Furthermore, they did not consider the inherent noise in skeleton data.
\section{Review of Denoising Autoencoders}%
\begin{figure}[tb]%
\centering%
\includegraphics[scale=0.75]{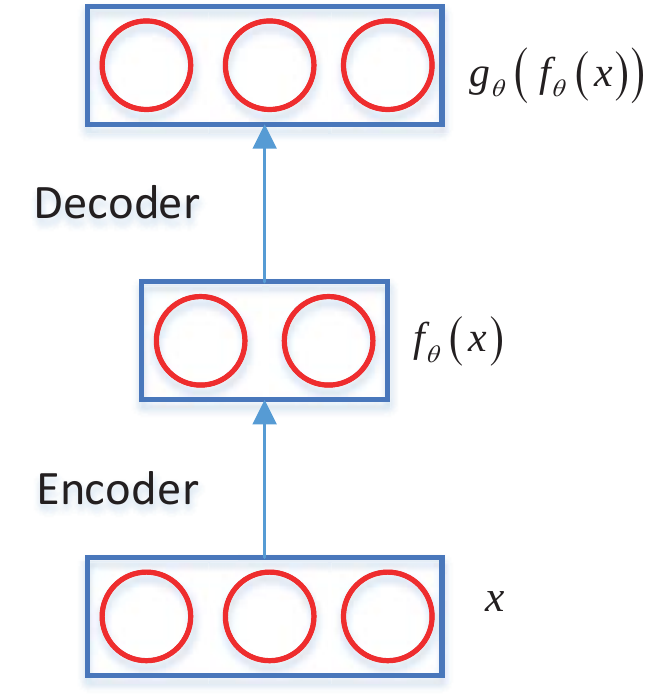}%
\caption{Single-layer $(AE)^{2}$}%
\label{fig:2}%
\end{figure}

\label{sec:3}%
Denoising Autoencoders have been to be very promising techniques for unsupervised feature learning~\cite{wu2015discriminative,wu2018autoencoder,hong2015multimodal}.
They have the strong ability of learning a compact and discriminative representation from the input with noises, while retaining the most important information.
And the representation is expected to be able to denoise and reconstruct the original input. 
With this characteristic, we hypothesis that Denoising Autoencoders is suitable for skeleton representation as the estimated positions of skeleton joints are likely not stable in practical applications. 
Here we discuss the framework of denoising autoencoders and its terminology based on the traditional \mbox{autoencoder $(AE)^{2}$~\cite{vincent2010stacked}}.
Figure~\ref{fig:2} shows a \mbox{single-layer $(AE)^{2}$}, which comprises two parts: encoder and decoder.%
\begin{sloppypar}%
\textbf{Encoder:} The deterministic \mbox{mapping~$f_{\theta}:{\mathbb R}^d\mapsto{\mathbb R}^{d'}$} transforms a \mbox{$d$-dimensional} input \mbox{vector~$x$} into a \mbox{$d'$-dimensional} feature \mbox{vector~$h$} in the hidden layer.
The typical form  of such mapping is an affine transformation followed by a non-linear \mbox{function~$s$}, which usually is a sigmoid \mbox{function $s(x)=\frac{1}{1+e^{-x}}$} or a hyperbolic tangent function \mbox{($s(x)=\frac{e^{x}-e{-x}}{e^{x}+e^{-x}}$)}.
For a \mbox{data set $\{x^{(1)},x^{(2)},\ldots, x^{(n)}\}$}, each \mbox{sample~$x^{(i)}$} can be defined:%
\end{sloppypar}%
\begin{equation}%
f_{\theta}(x^{(i)})=s(Wx^{(i)}+b).%
\label{eq:encode}%
\end{equation}%

\textbf{Decoder:} Based on the resulting hidden \mbox{representation~$h$}, a mapping function  $g_{\theta}:{\mathbb R}^{d'}\mapsto{\mathbb R}^{d}$ is further used to map it back to a reconstructed \mbox{$d$-dimensional} \mbox{vector~$r$} of the input space.
Similar to the encoder function $f_{\theta}$,  the typical form of $g_{\theta}$ is again an affine transformation optionally followed by a squashing non-linearity, that is,%
\begin{equation}%
g_{\theta}(h^{(i)})=s(W'h^{(i)}+c).%
\label{eq:dcode}%
\end{equation}%

In this model, the parameter set 
 \mbox{is~$\theta=\{W,W',b,c\}$}, \mbox{where $W$} \mbox{and $W'$} are the encoder and decoder weight matrices \mbox{and $b$} \mbox{and $c$} are the offset vectors \mbox{of $d'$} \mbox{and $d$} with corresponding dimension. 
 The \mbox{transpose of~$W'$} and \mbox{of~$W$} have  the same size, i.e., \mbox{is~$d'\times d$}.
%
The aim of the basic autoencoder is to minimize the reconstruction error.
Therefor, the objective function of the model is:%
\begin{equation}
L_{AE}(\theta)=\sum_{i} L\left(x^{(i)},g_{\theta}(f_{\theta}(x^{(i)}))\right).%
\label{eq:error1}%
\end{equation}%
Usually, the loss \mbox{function~$L$} is the square loss \mbox{function $L(x,r)=\left\|x-r\right\|^{2}$}.

\begin{figure}[tb]%
\centering%
\includegraphics[width=0.9\linewidth]{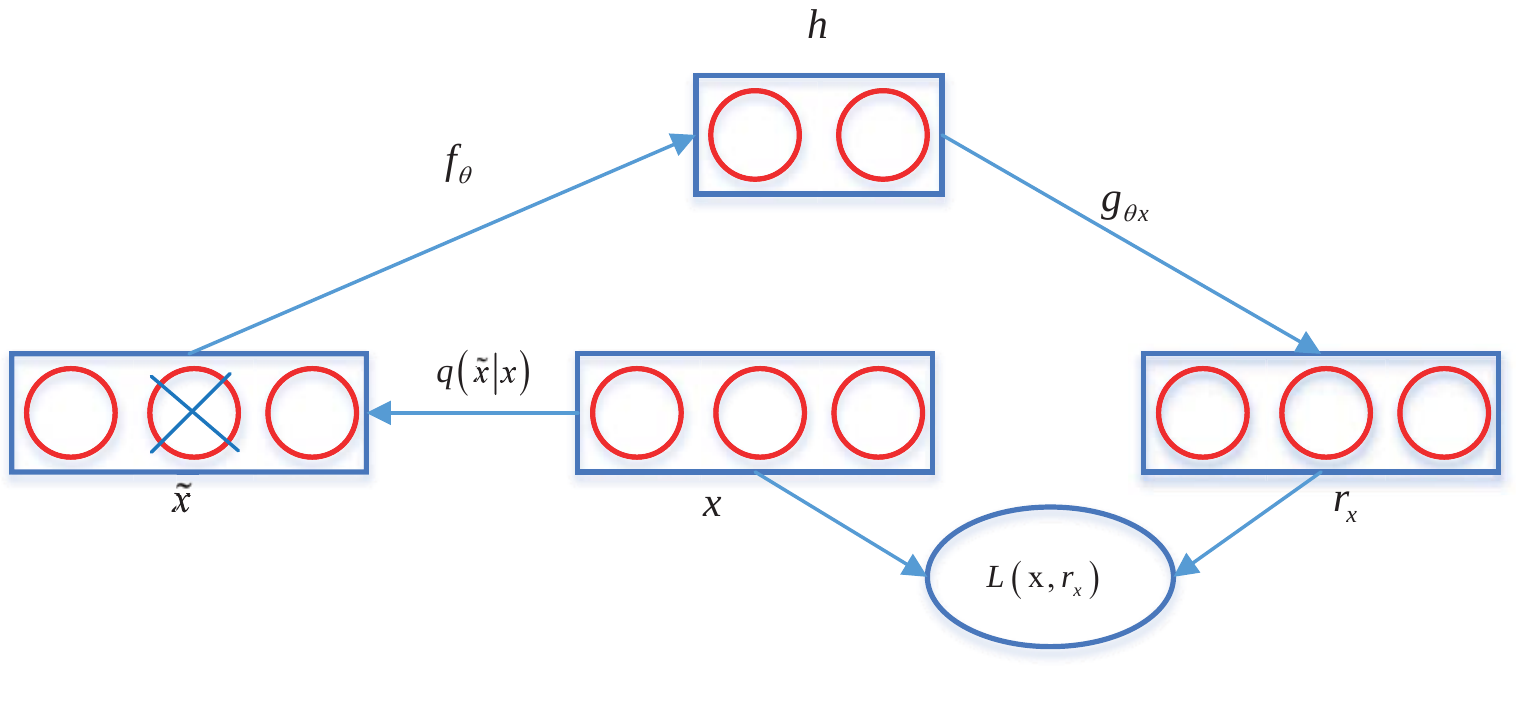}%
\caption{Architecture of the denoising autoencoder.}%
\label{fig:a}%
\end{figure}

In order to capture the joint distribution of the inputs and explore a strategy to denoise a corrupted version, Vincent et al.~\cite{vincent2010stacked} propose the Denoising Autoencoder~(DAE).
The DAE is like an ordinary autoencoder, with the difference that during learning, the input seen by the autoencoder is not the raw input but a stochastically corrupted version, as sketched in Figure~\ref{fig:a}.
The DAE is thus trained to reconstruct the original input from the noisy version.
Its input is the corrupted version $\tilde{x}$ of the raw data  \mbox{input~$x$}.
Usually, $\tilde{x}$ is generated using a stochastic \mbox{mapping~$\tilde{x} \sim q(\tilde{x}|x)$}.
The mapping corresponds to bilnding some elements that are randomly choosed from the input vector.
Here, it has a \mbox{probability~$q$} to be tuned to $0$ for each node in the input layer of the DAE. 
The objective function is normally as follow:%
%
\begin{equation}%
L_{DAE}(\theta)=\sum_{i}\mathbb{E}_{q(\tilde{x}|x)}\left[L\left(x^{(i)},g_{\theta}(f_{\theta}(x^{(i)}))\right)\right],
\end{equation}%
where \mbox{$\mathbb{E}_{q(\tilde{x}|x)}[\cdot]$ is} the expectation over the corrupted \mbox{examples~$\tilde{x}$} drawn from the corruption \mbox{process~$q(\tilde{x}|x)$}.%

Why the DAE can learn a compact representation from the corrupted data?
Actually, the process of the learning consists of reconstruction and denoising.
The process of denoising, that is, mapping a corrupted example back to an non-corrupted one, can be given an intuitive geometric interpretation under the manifold assumption~\cite{vincent2010stacked,chapelle2006semi}.
This assumption states that natural high dimensional data concentrates close to a non-linear low-dimensional manifold, yet most of the time the corruption vector is orthogonal to the manifold.
The DAE model can exactly denoise the input by learning to project the corrupted data back onto the low-dimensional manifold.%
\section{Proposed Skeleton Representation}%
\label{sec:4}%
Based on the DAE model, We now extend it by adding privileged information, that is
action category and skeleton temporal coordinate, to make the learned features more discriminative.
In addition, we present our novel stacking architecture for building a deep neural network.%
\subsection{Adding Action Category and Skeleton Temporal Coordinate}%
\label{sec:constrants}%
An action is a dynamical time sequence and human action recognition therefore has a temporal character.
Due to the subjective characteristics of the actor, there are subtle differences in the operation modes of actions.
Furthermore, actions in different categories may be very similar or have overlapping motions.
We argue that temporal information and action category are two important characteristics for action recognition.
Although DAE is very expressive because of its strong ability of capturing the joint distribution of the inputs, it learns representations without the temporal and action category information.
With this in mind, we modify the DAE by adding the privileged information of action category and skeleton temporary coordinate, as the constraint terms of the modified model, to make the model capable of emphasizing the disparities of different actions and the skeletons in them.%
\begin{figure}[tb]%
\centering%
\includegraphics[width=0.977\linewidth]{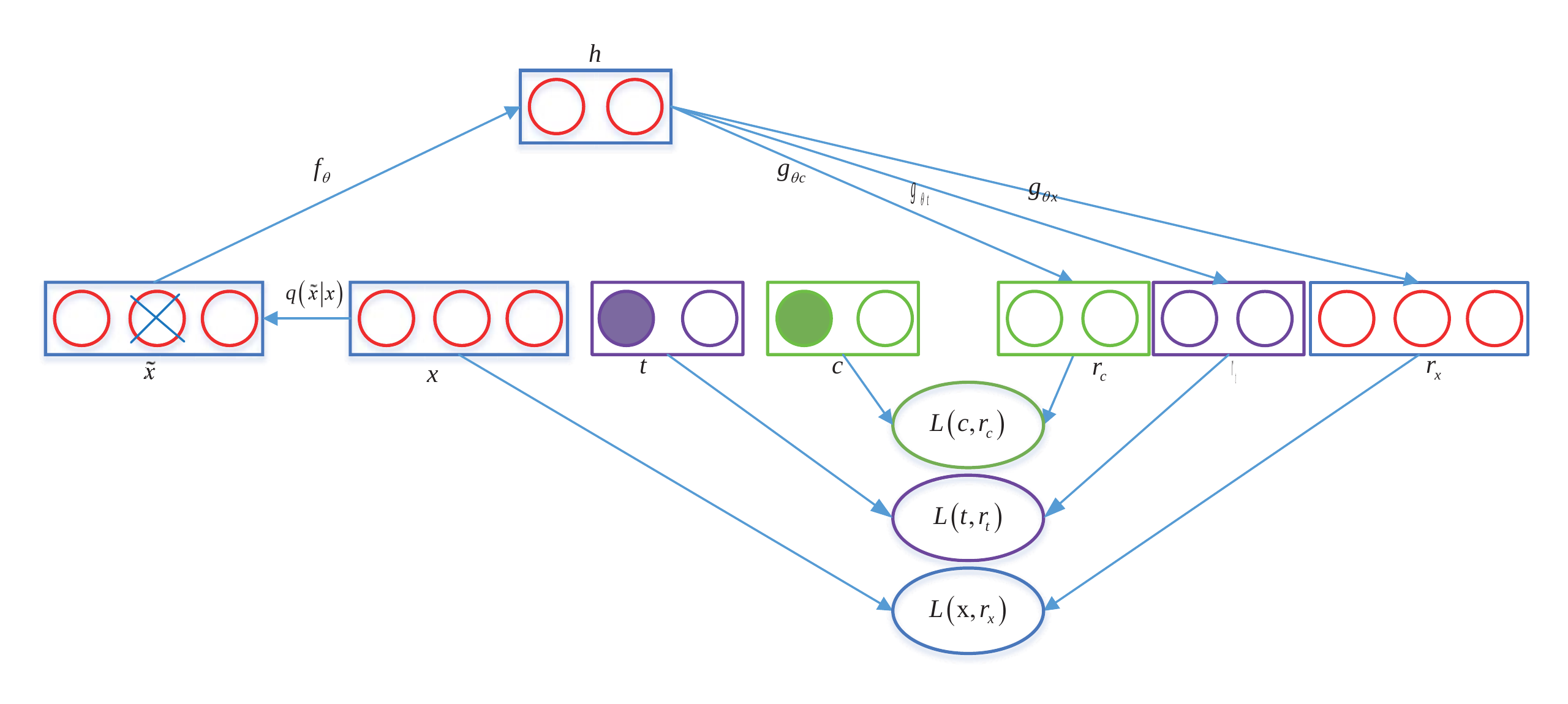}%
\caption{Architecture of DAE\_CCT}%
\label{fig:4}%
\end{figure}%
\begin{sloppypar}%
We name the modified model Denoising Autoencoder with Temporal and Categorical Constraints (DAE\_CTC), the architecture of the DAE\_CTC is illustrated in Figure~\ref{fig:4}.
The DAE\_CTC consists of three autoencoder parts: the skeleton autoencoder, the action category autoencoder, and the sekeleton temporal coordinate autoencoder. 
Compared to the structure of the usual~DAE, the~DAE\_CTC additionally has two objective vectors, $c$~and~$t$, which are used to indicate the action category and the relative temporal location of the input skeleton in the action.%
\end{sloppypar}%
The objective vector~$c$ is a vector of values~0 and~1, and its length equals the number of action categories.
Here we assume a total of~$l$ categories i.e., $|c|=l$.
In this vector, only one element has a non-zero value, and the others are all zero.
The index of the non-zero one indicates the action category that the current input belongs to.
During the training of the DAE\_CTC, the input category vector~$c$ is reconstructed using the hidden layer \mbox{function $g_{\theta c}$}, i.e., \mbox{$r_{c}=g_{\theta c}(h)$}.%
\begin{sloppypar}%
The sekeleton temporal autoencoder is designed for two aspects.
First, durations of the action sequences within the different category, even the same action, are likely different.
The skeleton temporal autoencoder is used to normalize speed variations and create weak alignments of these sequences.
On the other hand, this autoencoder somewhat simulates the Short-Term Memory of the human cognition system~\cite{REF45}.
In the 1950s, George Armitage Miller conducted a series of quantitative analyses with respect to the short-term memory capacity.
A famous observation that the short-term memory capacity of the human can hold approximately seven chunks~\cite{REF46,REF47} is concluded base on such experiments.
With this observation, we decided to artificially divide each instance into seven sections using uniform sampling.
We define a vector~$t$ of seven elements, whose single positive element indicates the relative temporal position of the current skeleton.
The input temporal vector~$t$ is reconstructed using the hidden layer \mbox{function~$g_{\theta t}$}, i.e., \mbox{$r_{t}=g_{\theta t}(h)$}.%
\end{sloppypar}%
The reconstruction vector \mbox{of~$x$} \mbox{us~$r_x$}, obtained from the \mbox{mapping~$g_{{\theta}x}$}.
The new training objective of \mbox{DAE\_CCT} is defined in Equation~\ref{eqtrain}.
\begin{equation}%
\begin{aligned}%
&L_{DAE\_CCT}(\theta)=\sum_{i}\mathbb{E}_{q(\tilde{x}|x)}\\
&\left[L(x^{(i)},g_{{\theta}x}(f_{\theta}(x^{(i)})))+{\lambda}L(c^{(i)},g_{{\theta}c}(f_{\theta}(c^{(i)})))+{\beta}L(t^{(i)},g_{{\theta}t}(f_{\theta}(t^{(i)})))\right]%
\label{eqtrain}%
\end{aligned}%
\end{equation}%
Here, \mbox{$\lambda$ and $\beta$} are the hyper-parameters controlling the constraint strength of the action category and the skeleton temporal, respectively.
By this, it can establish a trade-off between reconstructing the input data and preserving action category information as well as temporal information.
Similar to the training of the DAE, we use the stochastic gradient descent algorithm to optimize the DAC\_CCT.
%
%
%
\subsection{Stacked Architecture}%
\begin{figure}[tb]%
\centering%
\includegraphics[width=0.977\linewidth]{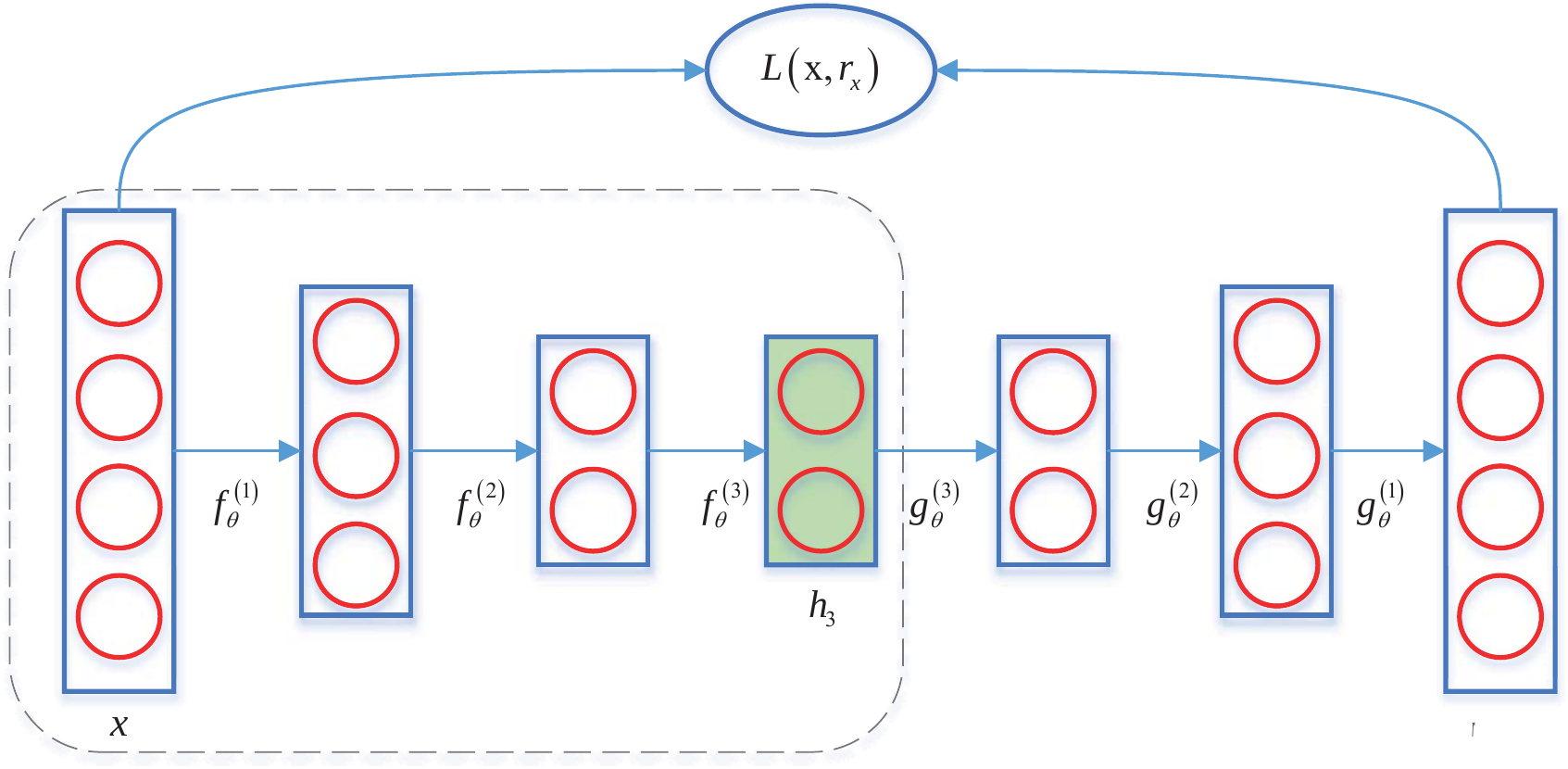}%
\caption{Fine tuning of the stacking architecture}%
\label{fig:b}%
\end{figure}
By stacking several layers of \mbox{DAE\_CCT}, we can build an enhanced deep neural network, as shown in Figure~\ref{fig:b}.
We use a greedy layer-wise training strategy:
We train the first layer as shown in Figure~\ref{fig:b} and obtain the encoding \mbox{function~$f_{{\theta}1}$}, which is used to compute the output.
The output is then used to train the second layer autoencoder to obtain \mbox{function~$f_{{\theta}2}$}.
This process is applied iteratively.
Finally, the output of the last autoencoder is the output for the stacked architecture.

%
%
\section{Temporal Registration and Classification}%
\label{sec:5}
In this section, we propose a new method of temporal registration to mitigate the nuisance variation of temporal misalignments.
We represent the dynamic features of a whole action sequence using the recently proposed Fourier temporal pyramid representation~\cite{wang2012mining}, and perform action recognition with a \mbox{one-vs-all} linear SVM.%
\subsection{Temporal registration method}%
\begin{sloppypar}%
An action is represented as a sequence of skeleton-level features \mbox{$H = (h_{1}, \dots, h_{T})$}, where \mbox{$h_{i}$ is} the \mbox{$i$\textsuperscript{th} skeleton} descriptor learned with \mbox{DAE\_CCT}.
The details of such features were discussed in \mbox{Section~\ref{sec:3}}.
Here, $T$~is the reference length of actions.
An action data set then is specified as \mbox{$\{(H_{1}, c_{1}), (H_{2}, c_{2}), \dots, (H_{N}, c_{N})\}$}, \mbox{where $H_{i}$} is an action sequence \mbox{and $c_{i}$} is its action category label.
We assume \mbox{$l$ categories}, i.e., \mbox{$C = \{1, \dots, l\}$}.
For each action class, we define a phantom action template~$P$ which consists of a sequence of atomic actions:%
\end{sloppypar}%
\begin{equation}%
\label{eq:min_template}%
P = \{p_{1},p_{2}, \dots, p_{T}\}%
\end{equation}%
Here, \mbox{$p_{j}$ denotes} the frame-level features for the \mbox{$j$\textsuperscript{th} atomic} action.

\begin{figure}[tb]%
\centering%
\includegraphics[width=0.9\linewidth]{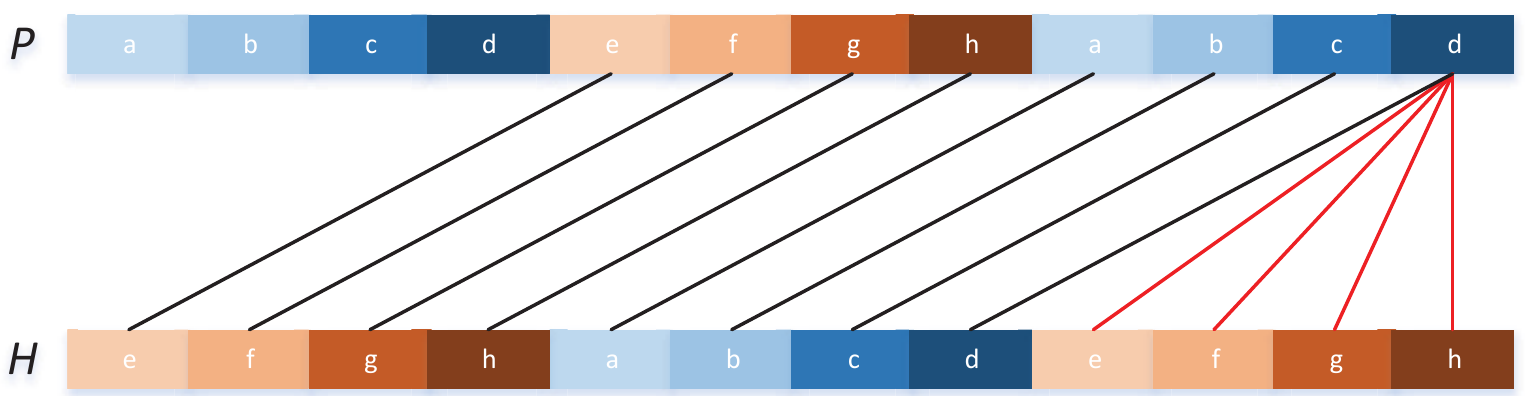}%
\caption{Warping path between periodic actions using DTW}%
\label{fig:c}%
\end{figure}

Dynamic Time Warping (DTW) is a frequently-used method~\cite{vemulapalli2014human, amor2016action} for action registration.
When registering a action sequence using DTW, the initial point and end point of the warping path are fixed, and the path only flows forwards, i.e., cannot go back in time.
Existing works that use DTW to warp a training sequence to a phantom action are likely to produce some temporal misalignments, yielding poor performance especially for periodic actions (such as \emph{``waving''}).
The red lines in  Figure~\ref{fig:c} indicate the misalignments.
We denote the phantom template \mbox{with $P$} and an intra-class action sequence \mbox{as $H$}.
To mitigate this nuisance, we perform registration without the rigorous restriction along the warping flow and locally register the warping sequence.%

\begin{figure}[tb]%
\centering%
\includegraphics[width=0.9\linewidth]{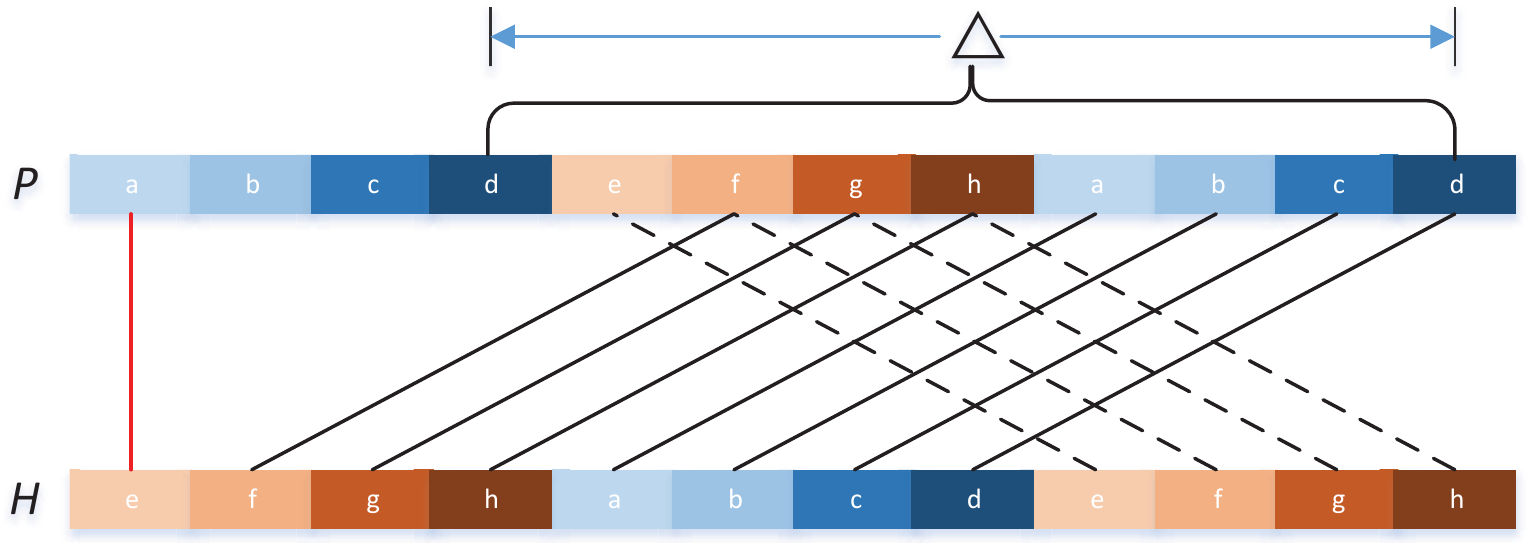}%
\caption{Warping path between intra sequences using LRWS\_intra}%
\label{fig:d}%
\end{figure}

An action instance is composed of multiple simple and relative sub-actions.
Thus, when performing registration between a phantom template and an intra-class sequence, we can first locate a warping region and then perform registration in the local region.
We call this \mbox{process LRWS\_intra}, which is defined in  Equation~\ref{eq:dtw-warping}, where \mbox{$\Delta$ is} the so-called local warping region.%
\begin{equation}%
\label{eq:dtw-warping}%
\mathit{LRWS\_{intra}}(P,H,i) = \arg\min_{j\in \Delta(i)}||P(j)-H(i)||_{2}^{2}.%
\end{equation}%
When performing registration between the aforementioned two sequences \mbox{using LRWS\_intra}, we can obtain a new warping path, as shown in Figure~\ref{fig:d}.
Notice that fewer misalignments are produced than with DTW.
Differing from the DTW, the warping paths of \mbox{the LRWS\_intra} does not rigorously impose the flowing direction, which means that \mbox{LRWS\_intra permits} local disorder, as illustrated with the dotted lines.%

\begin{figure}[tb]%
\centering%
\includegraphics[width=0.9\linewidth]{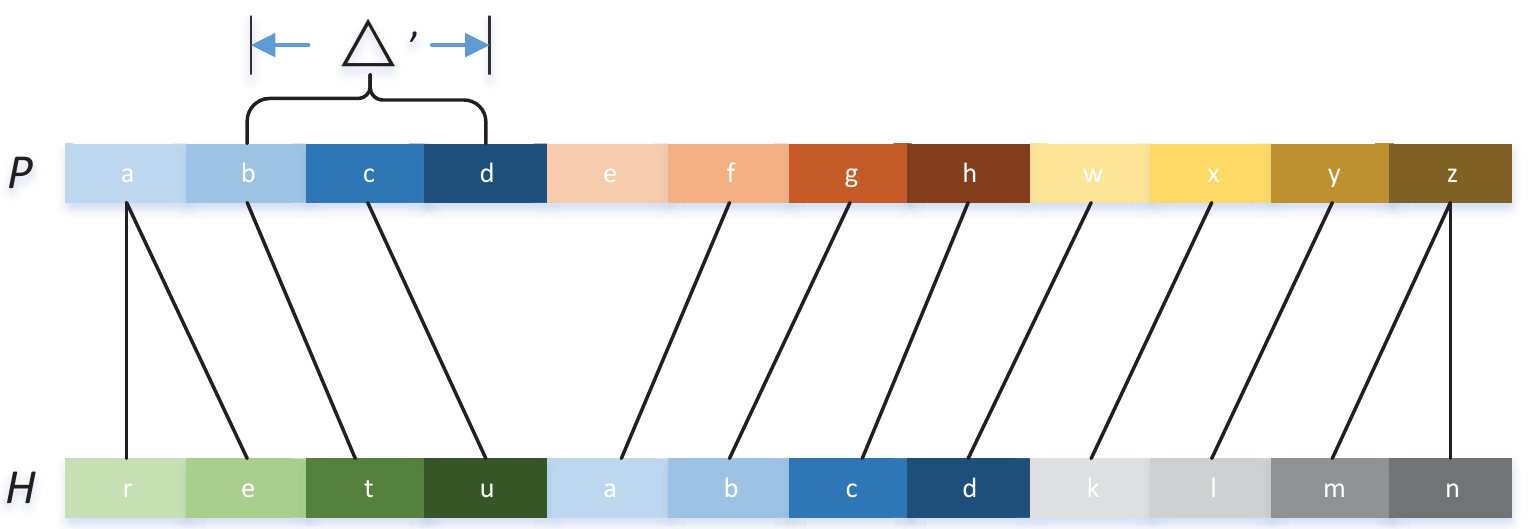}%
\caption{Warping path between the inter-action using LRWS\_inter}%
\label{fig:inter}%
\end{figure}

\mbox{LRWS\_intra allows} the registration of periodic sequences.
However, it can only handle \mbox{intra-actions} and cannot register \mbox{inter-actions}.
In temporal registration, we focus on the joint task \mbox{of inter-class} \mbox{and intra-class} registration.
We argue that a better phantom action template should depend on not only the minimized \mbox{intra-class differences} but also on the \mbox{inter-class disparities}.
We define a second temporal registration process, \mbox{named LRWS\_inter} in Equation~\ref{eq:max_template}, in order to improve the discriminative ability of the phantoms.
As shown in Figure~\ref{fig:inter}, we \mbox{register~$h'_{i}$} within a local region of size~$\Delta'$ around the atomic \mbox{action~$p_{i}$} by maximizing their difference along the inter-class sequence~$H'$.%
\begin{equation} \label{eq:max_template}%
\mathit{LRWS\_{inter}}(P,H',i) = \arg\max_{j\in \Delta'(i)}||P(j)-H'(i)||_{2}^{2}%
\end{equation}%
\begin{algorithm}%
\caption{Calculating the $n$\textsuperscript{th} category phantom}%
\label{algo:1}%
\small{%
\begin{algorithmic}[1]%
\Require The training sequences:
\State Maximum number of \mbox{iterations: $max$;} \mbox{threshold: $\zeta$;} weighing \mbox{coefficient: $\eta$}, \mbox{$0 \leq \eta \leq 1$;}
\Ensure \mbox{$n$\textsuperscript{th} category} phantom action \mbox{template $P\{p_{1},\dots,p_{L}\}$;}
\State Initialize: $P = H_{1}^{n}, iter =0$;
\For{$iter \in 1{\dots}max$}
\State Register each sequence in the $n$\textsuperscript{th} category to phantom template~$P$ using $\mathit{LRWS\_{intra}}$, and get a set of warped \mbox{sequences $W\{w_{1},\dots,w_{N_{n}}\}$;}
\State Separately select a random sequence from each category of the rest to make up an inter-class \mbox{set $H^{0}$,} which contains \mbox{$l-1$ sequences;}
\State Register each sequence \mbox{in $H^{0}$} to the phantom \mbox{template $P$} \mbox{using $\mathit{LRWS\_{inter}}$} and get a set of warped \mbox{sequences $W'\{w_{1}',\dots,w_{l-1}'\}$;}
\State Compute a new \mbox{phantom $P'$} using
\State $P'= (1-\eta)\frac{1}{N_{n}}\sum_{j=1}^{N_{n}}W(j) + \eta\frac{1}{l-1}\sum_{i=1}^{l-1}W'(i)$;
\If{$||P'-P||_{2}^{2} \leq \zeta$}
\State break;
\Else
\State $P=P'$;
\EndIf
\EndFor
\end{algorithmic}}%
\end{algorithm}%

Generating a phantom template for a specified action category~$P_{n}$ essentially means combining \mbox{LRWS\_intra} and \mbox{LRWS\_inter} with $k$-means clustering, where \mbox{$k$ is 1}.
We take the $n$\textsuperscript{th} category phantom as a case and describe this process in Algorithm~\ref{algo:1}.
In each iteration, we randomly select an action sequence from each category of the rest to make up an \mbox{inter-class set}.
The action sequence which belonged to the \mbox{$n$\textsuperscript{th} category} is registered \mbox{using LRWS\_intra}, and those from the other classes are registered \mbox{using LRWS\_inter}.
Then, by weighting these two kinds of warped sequences, we yield a new phantom template in each iteration.
We will obtain the \mbox{$n$\textsuperscript{th} category} phantom template when the iteration stops.%
\subsection{Fourier Temporal Pyramid Representation}%
\label{sec:ftp}%
The Fourier temporal pyramid~(FTP) representation~\cite{wang2012mining} is used to capture the temporal structure of an action.
Motivated by Spatial Pyramid Matching~(SPM)~\cite{lazebnik2006beyond}, a Short-Time Fourier Transform is applied for each dimension of a warped instance sequence.
We recursively partition the sequence into increasingly finer segments along the temporal direction.
Typically, three levels, each \mbox{containing 1}, 2, \mbox{and 4} segments, respectively, are used.
The final feature vector is the concatenation of the Fourier coefficients from all the segments.
We perform action recognition by classifying these final feature vectors by using a \mbox{one-vs-all} linear SVM~\cite{chang2011libsvm}.
This approach is sketched in Figure~\ref{fig:1}.%
\section{Experimental Evaluation}%
\label{sec:6}%
\begin{sloppypar}%
In this section, we evaluate our approach and compare with three other skeletal representations and several recent works on the benchmark data sets \mbox{MSR-Action3D~\cite{li2010action}}, \mbox{UTKinect-Action~\cite{xia2012view}}, and \mbox{Florence3D-Action~\cite{seidenari2013recognizing}}.%
\end{sloppypar}%

For all the skeletons in the three data sets, we normalize them to the unified coordinate system with a origin of the hip center. 
In order to make the skeletons view-invariant, similar to \cite{vemulapalli2014human}, we rotate both ends of the hip bone to parallel the global \mbox{$x$-axis}.
And for each data set, we select one referenced skeleton and normalize all other skeletons without changing their joint angles in order to make the skeletons scale-invariant.
The comparative results of all the state-of-the-art methods on the data sets are from their corresponding papers.%
\subsection{Comparative Skeletal Representations}%
\begin{sloppypar}%
Let \mbox{$S = (V;E)$ be} a skeleton, where \mbox{$V = (v_{1},\dots,v_{N})$ denotes} the joints and \mbox{$E = (e_{1},\dots, e_{M})$ denotes} the rigid body parts.
There are three simple and effective skeleton representation methods: the joint positions~(JP), the pairwise relative positions of the joints~(RJP), and the joint angles~(JA).
\end{sloppypar}%
\begin{itemize}%
\item{JP:} the concatenation of the \mbox{3D coordinates} of all the \mbox{joints $v_{1},\dots,v_{N}$.}%
\item{RJP:} the concatenation of all the vectors \mbox{$\overrightarrow{v_{i}v_{j}}, 1\leq i < j \leq N$.}%
\item{JA:} the concatenation of the quaternions corresponding to all joint angles.%
\end{itemize}%
%
%
Based on our proposed model, we can define three more variants: \mbox{\emph{DAE\_CC}}, which only retains the constraint of the action category, \mbox{\emph{DAE\_TC}}, which only retains the constraint of the temporal relationship, and \mbox{\emph{DAE}}, which uses neither action category information nor temporal constraints.

In order to verify the effectiveness of our approach, we compare it with the above six alternative methods in the experiments.
It should be noted that all representations use the same temporal modeling and classification described in Section~\ref{sec:ftp}.%
\subsection{Evaluation Settings}%
We follow the cross-subject test setting from~\cite{li2010action} \mbox{for MSR-Action3D}.
There, the data set was divided into \mbox{subsets $AS_1$,} $AS_2$, \mbox{and $AS_3$,} each consisting of \mbox{8 actions}.
We perform recognition on each subset separately. %
%
%
For \mbox{UTKinect-Action,} we follow the experimental protocol from~\cite{zhu2013fusing}, where half of the subjects are used for training while the remaining for testing.
For \mbox{Florence3D-Action,} we apply the leave-one-subject-out experimental protocol proposed \mbox{in \cite{seidenari2013recognizing}}.

For all the experiments, we train a deep architecture with a three-stacked DAE\_CCT .
The hidden node setting of this architecture is given in Table~\ref{tab:2}.
The action category and temporal constraint parameters are both experimentally assigned \mbox{to 1.5.}
We penalize the average \mbox{output $\bar{h}{_j}$} of the second autoencoder, pushing it \mbox{to 0.1,} in order to add sparsity and learn an over-completed representation of a skeleton.
In the phase of temporal registration, we take the length of a chunk as the size of the local warping region, which is consistent with the short-term memory capacity discussed in Section~\ref{sec:constrants}.
Finally, a FTM over three levels, each \mbox{containing 1,} 2, and \mbox{4 segments,} is used to model the temporal relationships.%
\begin{table}[tb]%
\centering%
\caption{Hidden node setting of the stacked DAE\_CCT}%
\label{tab:2}%
\small{%
\begin{tabular}{|c|c|c|c|}
\hline
Layer& MSR Action3D & UTKinect-Action & Florence3D-Action\\
\hline
$H_{1}$&200&150&100\\
\hline
$H_{2}$&400&300&200\\
\hline
$H_{3}$&800&600&400\\
\hline
\end{tabular}%
}%
\end{table}%
\subsection{Experimental Results and Analysis}%
\subsubsection{Comparison with the alternative approaches}%
We compare the \mbox{DAE\_CTC} with the alternative approaches JP, RJP, JA, DAE, DAE\_CC, and DAE\_TC on \mbox{MSR-Action3D} using the protocol of~\cite{li2010action}.
The experimental results are shown in Table~\ref{tab:3}.
We can see that \mbox{DAE\_CTC} achieves the best average accuracy and outperforms the other six methods.
Its independent recognition rates of the each subsets are the best as well.
Furthermore, we notice that our approach performs consistently well on the three subsets, which indicates that \mbox{DAE\_CTC} is more robust.
We also find that the performance of the three simplified versions of \mbox{DAE\_CTC} is promising.
The fact that \mbox{DAE\_CCT} obtains higher average accuracy than the other three variants proves the significance of constraints of action category and temporal relationships.%
\begin{table}[tb]%
\centering%
\caption{Recognition rates on MSR-Action3D}%
\label{tab:3}%
\small{%
\begin{tabular}{|c|c|c|c|c|c|c|c|}
\hline
Dateset & JP   & RJP   & JA    & DAE   & DAE\_CC & DAE\_TC  &DAE\_CTC\\
\hline
$AS_1$ & 90.83 & 90.41 & 89.23 & 91.68 & 93.37    &92.56     & \textbf{95.00}\\
\hline
$AS_2$ & 81.75 & 82.72 & 78.36 &86.46  & 88.92    &88.20     & \textbf{90.99}\\
\hline
$AS_3$ & 93.82 & 93.54 & 92.18 &94.28  & 95.65    & 94.90    & \textbf{96.13}\\
\hline
Average & 88.80 & 88.89 & 86.59 &90.64 & 92.65    & 91.89    & \textbf{94.04}\\
\hline
\end{tabular}%
}%
\end{table}%

\begin{figure}[tb]%
\centering%
\strut\hfill\strut%
\subfloat[$AS_1$]{\includegraphics[width=0.435\linewidth]{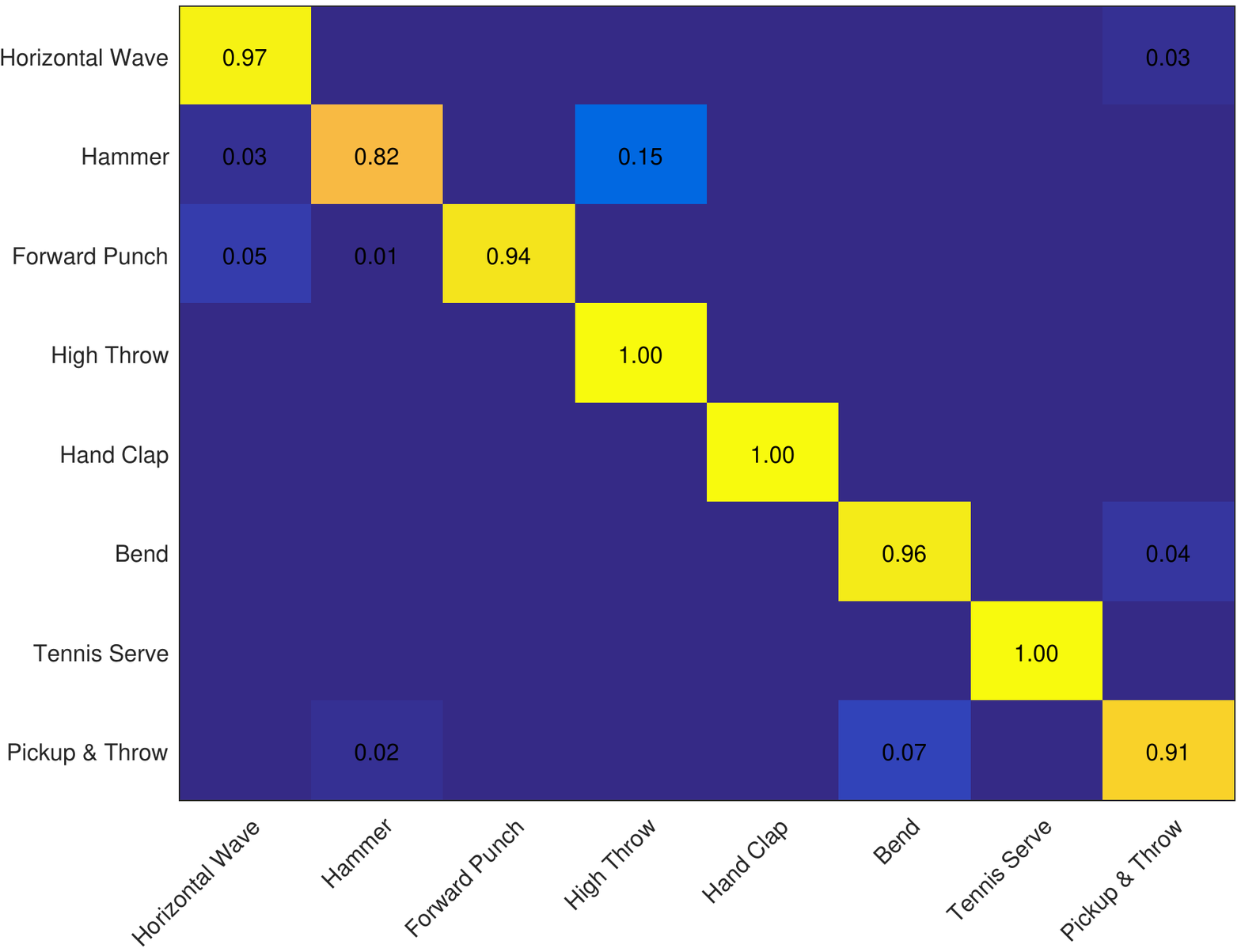}}%
\strut\hfill\strut%
\subfloat[$AS_2$]{\includegraphics[width=0.435\linewidth]{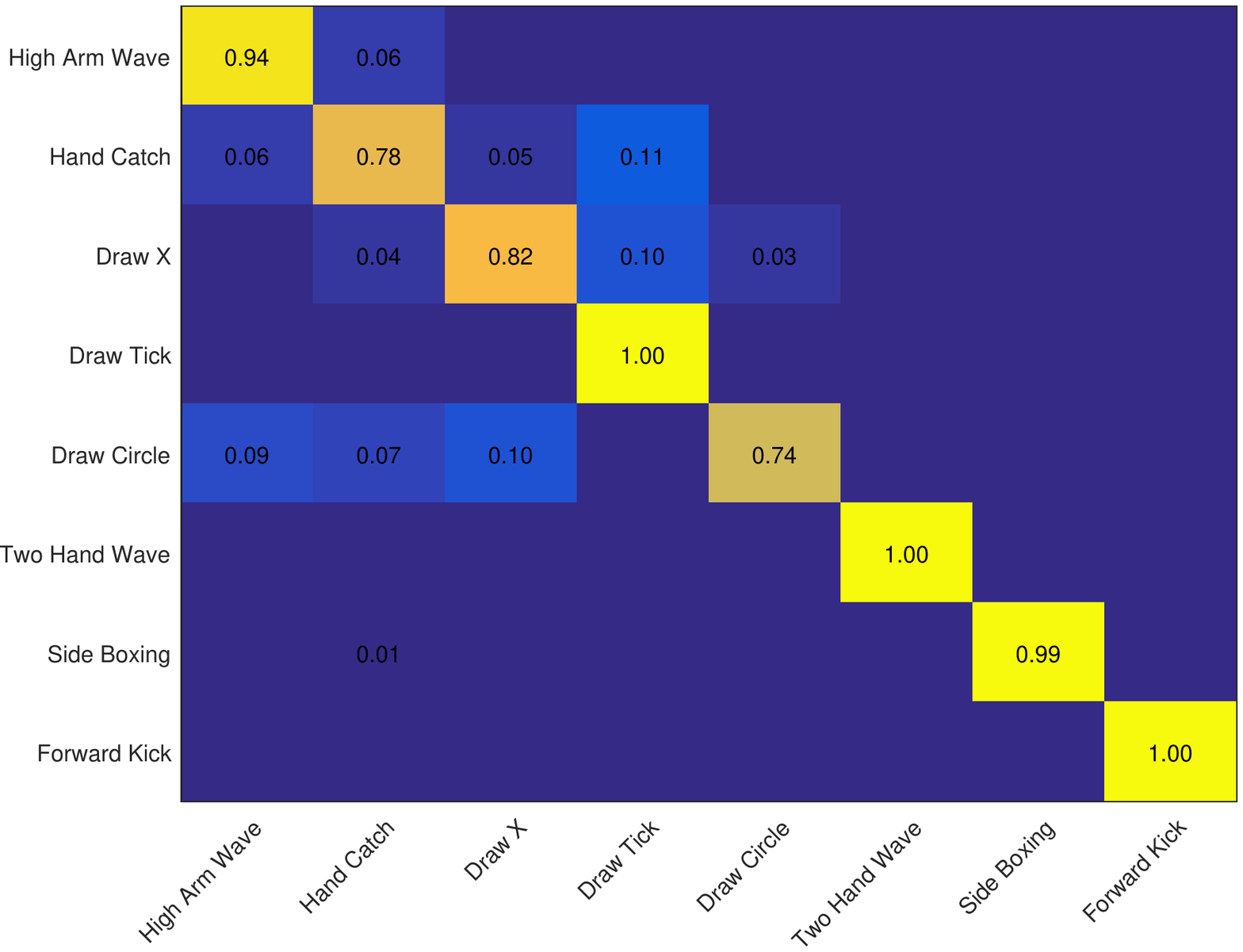}}%
\strut\hfill\strut%
\\\noindent%
\strut\hfill\strut%
\subfloat[$AS_2$]{\includegraphics[width=0.435\linewidth]{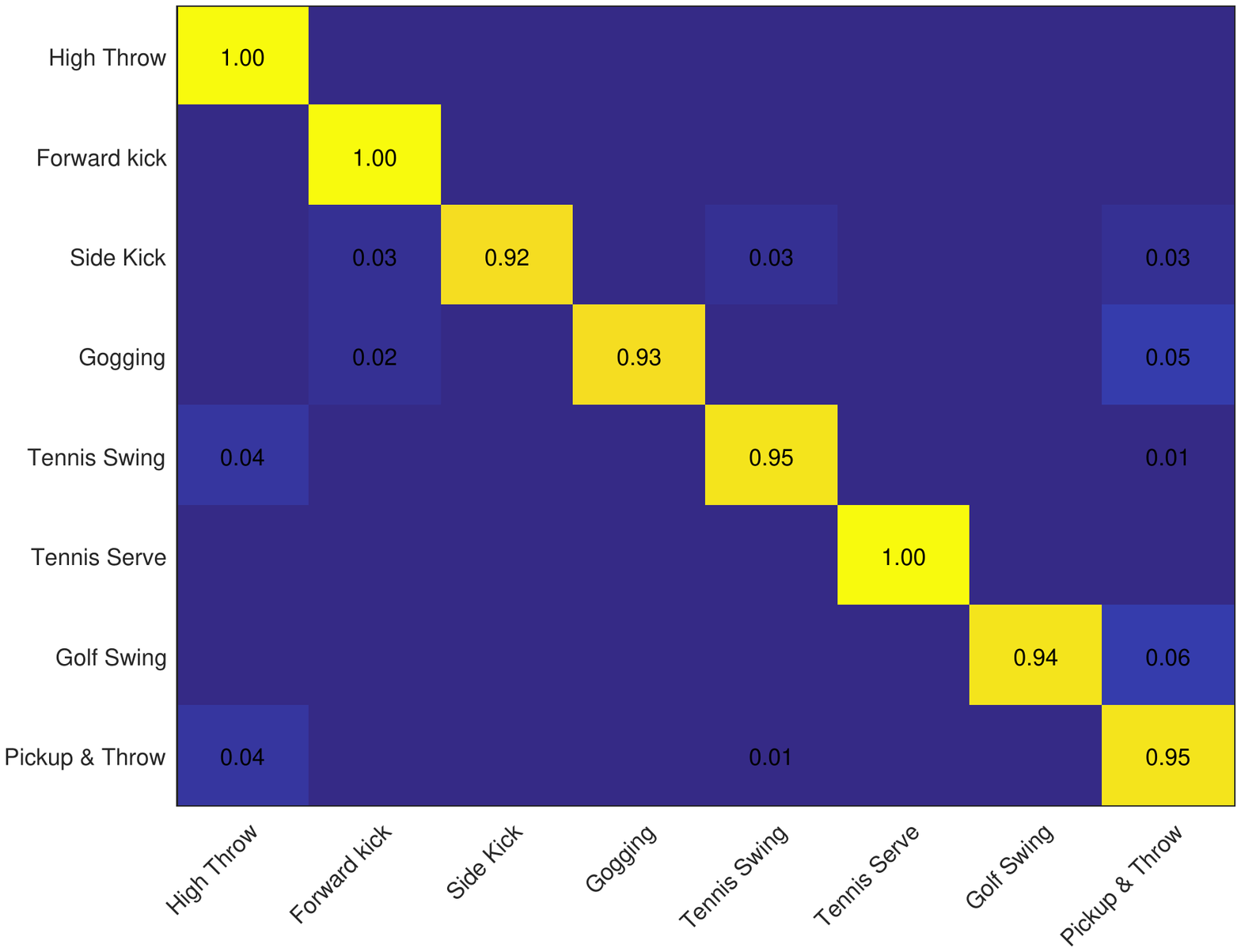}}%
\strut\hfill\strut%
\caption{Confusion matrices of DAE\_CCT on MSR-Action3D for specific protocols.}%
\label{fig:AS_confusion}%
\end{figure}

The confusion matrices on the three action sets are shown in Figure~\ref{fig:AS_confusion}.
Misclassifications mainly occur among very similar actions.
For example, the action \emph{``hammer''} in the case \mbox{of~$AS_1$} is often misclassified to \emph{``high throw''}, while \emph{``pickup and throw''} is misclassified to \emph{``bend''}.
The actions \emph{``hammer''} and \emph{``high throw''} share a large overlap in the sequences.
The action \emph{``pickup and throw''} just has one more \emph{``throw''} move than \emph{``bend''} and the \emph{``throw''} move often holds few frames in the sequence. Distinguishing them is very challenging with the skeleton data only.%
\begin{sloppypar}%
Following the protocols of~\cite{zhu2013fusing, seidenari2013recognizing}, recognition rates for various skeletal representations on \mbox{UTKinect-Action} and \mbox{Florence3D-Action} are shown in Table~\ref{tab:4}.
\mbox{DAE\_CTC} has the best results on all the data sets.
The simplified versions of \mbox{DAE\_CTC} also obtain excellent results compared to the other three commonly-used representations.%
\end{sloppypar}%
\begin{table}[tb]%
\centering%
\caption{Recognition rates on UTKinect-Action}%
\label{tab:4}%
\resizebox{0.99\linewidth}{!}{\small{%
\begin{tabular}{|c|c|c|c|c|c|c|c|}
\hline
Dateset & JP & RJP & JA & DAE & DAE\_CC & DAE\_CT & DAE\_CCT\\
\hline
UTKinect-Action& 94.54 & 94.86 & 93.67 &95.07& 96.42 & 95.63 & \textbf{97.20}\\
Florence3D-Action& 80.13 &81.35 & 79.86 &82.51& 86.42 & 85.93 & \textbf{88.20}\\
\hline
\end{tabular}%
}}%
\end{table}

\begin{figure}[tb]%
\centering%
\strut\hfill\strut%
\subfloat[\emph{UTKinect-Action}]{\includegraphics[width=0.435\linewidth]{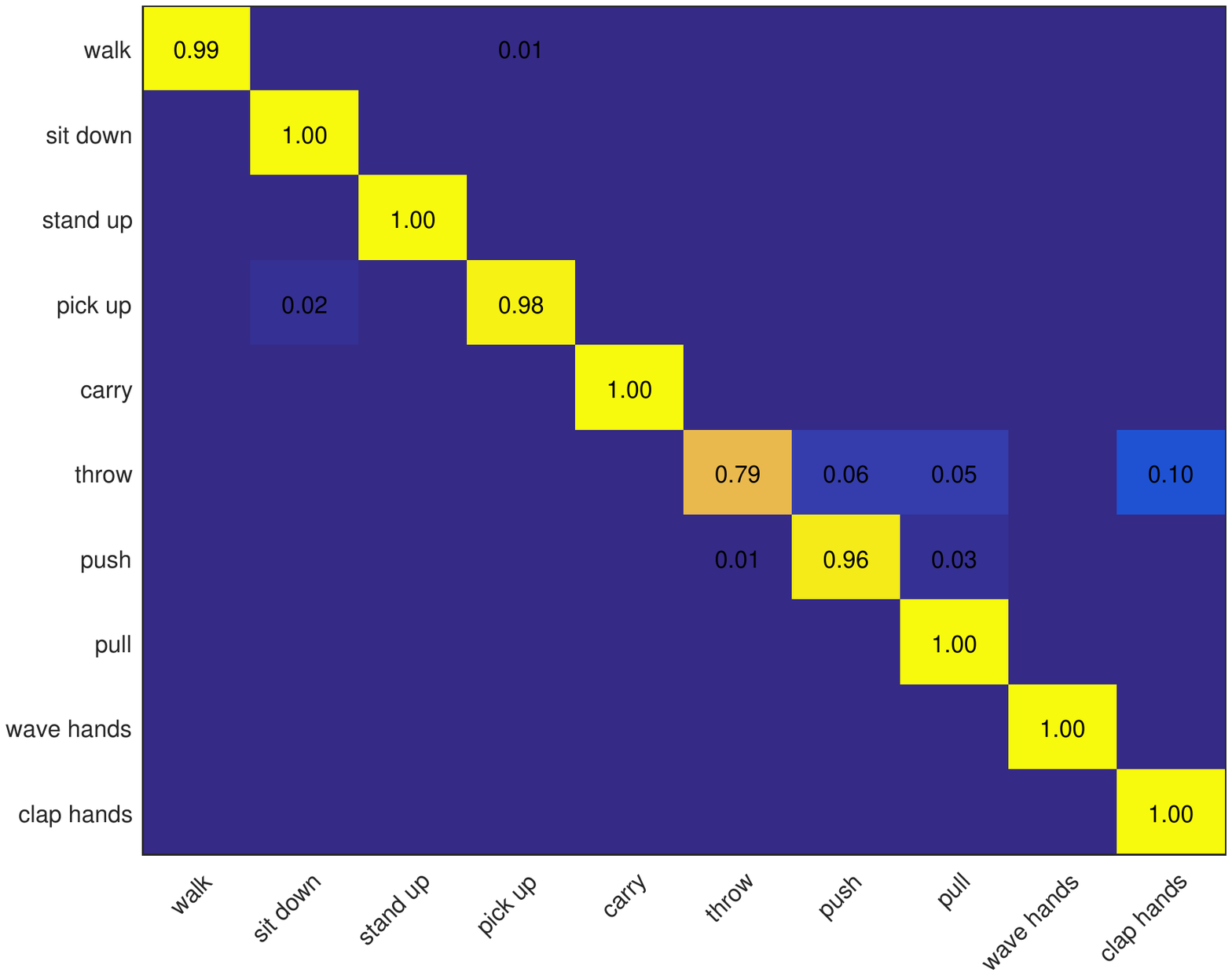}}%
\strut\hfill\strut%
\subfloat[\emph{Florence3D-Action}]{\includegraphics[width=0.435\linewidth]{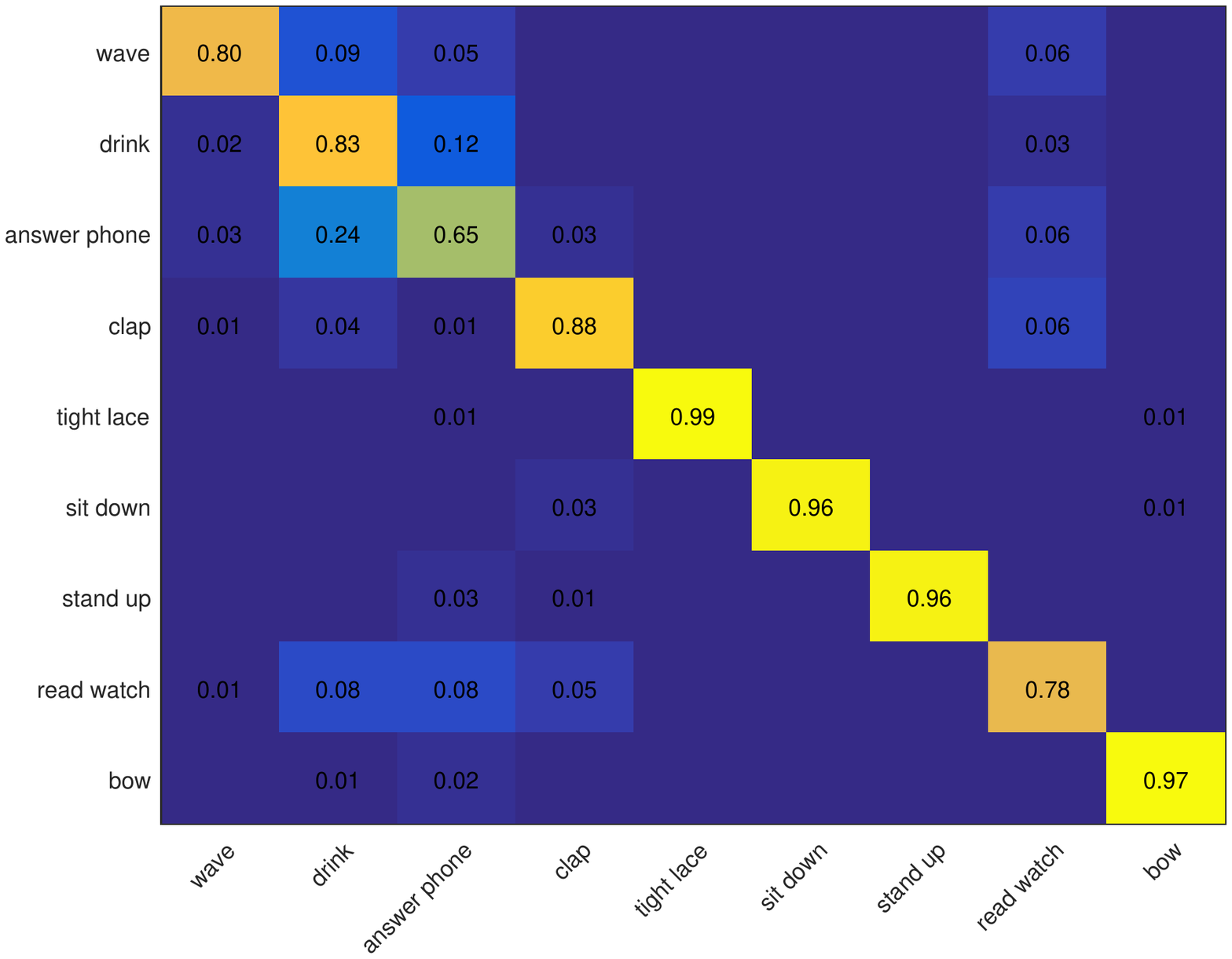}}%
\strut\hfill\strut%
\caption{Confusion matrices on UTKinect-Action and Florence3D-Action }%
\label{fig:12}%
\end{figure}

Figure~\ref{fig:12} illustrates the confusion matrices on the \mbox{UTKinect-Action} and \mbox{Florence3D-Action} data sets.
Our approach here performs very well on most of the actions.
Still, some actions, such as \emph{``throw''} and \emph{``push''} in \mbox{UTKinect-Action} or \emph{``drink''} and \emph{``answer phone''} in \mbox{Florence3D-Action}, are too similar to each other for our method to capture the differences.%
\begin{sloppypar}%
At this point, we have evaluated seven approaches including JP, RGP, JA, DAE, \mbox{DAE\_CC,} \mbox{DAE\_CT,} \mbox{and DAE\_CCT} on three benchmark data sets.
We can conclude that \mbox{DAE\_CCT} is a very effective approach for skeleton-based action recognition and the constraints of temporal relationship and category are significant improvements for skeleton representation.%
\end{sloppypar}%
\subsubsection{Comparison with state-of-the-art}%
We now compare the \mbox{DAE\_CCT} with various state-of-the-art approaches.
Table~\ref{tab:5} shows the comparison results on \mbox{MSR-Action3D} using the protocol of~\cite{li2010action}.
\mbox{DAE\_CCT} achieves an accuracy \mbox{of 94.04\%.}
It performs comparable to the state-of-the-art method given in~\cite{du2015hierarchical} and better than most other skeleton-based approaches.
The results of the three variants are also competitive.%
\begin{table}[tb]%
\centering%
\caption{Comparison with the state-of-the-art on \mbox{MSR-Action3D} using the protocol from~\cite{li2010action}}%
\label{tab:5}%
\small{%
\begin{tabular}{lllll}
\hline
Method & AS1 & AS2 & AS3 & Ave.\\
\hline
Li \emph{et al.}~\cite{li2010action} & 72.9& 71.9 &79.2& 74.7\\
Hussein \emph{et al.}~\cite{hussein2013human} & 88.04 & 89.29 & 94.29 & 90.53\\
Chen \emph{et al.}~\cite{chen2013real}  & 96.2 & 83.2 & 92.0 & 90.47\\
Gowayyed \emph{et al.}~\cite{gowayyed2013histogram} & 92.39 & 90.18 & 91.43 & 91.26\\
Vemulapalli \emph{et al.}~\cite{vemulapalli2014human} & 95.29 &83.87& 98.22 &92.46\\
Du \emph{et al.}~\cite{du2015hierarchical} &93.33&94.6&95.50&\textbf{94.49}\\
\hline
DAE&91.18&86.46&94.28&90.64\\
\hline
DAE\_CC&93.37&88.92&95.65&92.65\\
\hline
DAE\_TC&92.56&88.20&94.90&91.89\\
\textbf{DAE\_CTC}& 95.50& 90.99 &96.13& \textbf{94.04}\\
\hline
\end{tabular}
}%
\end{table}%

The comparison results on \mbox{UTKinect-Action} and \mbox{Florence3D-Action} are also given in Table~\ref{tab:state}.
Our approach achieves very good results on both data sets, where it can even reach the performance of the method given in~\cite{devanne20143}.%
\begin{table}[tb]%
\centering%
\caption{Comparison with the state-of-the-art on \mbox{UTKinect-Action} and \mbox{Florence3D-Action}}%
\label{tab:state}%
\small{%
\subfloat{%
\scalebox{0.95}{%
\begin{tabular}{|c|c|}%
\hline
\multicolumn{2}{|c|}{UTKinect-Action data set (protocol~\cite{zhu2013fusing})}\\
\hline
Random forests~\cite{zhu2013fusing} & 87.90\\
Points in a Lie Group~\cite{vemulapalli2014human} &97.08\\
\hline
DAE&95.07\\
DAE\_CC&96.42\\
DAE\_CT&95.63\\
\textbf{DAE\_CCT} &\textbf{97.20} \\
\hline
\multicolumn{2}{c}{~}\\[-2ex]
\hline
\multicolumn{2}{|c|}{Florence3D-Action data set (protocol~\cite{seidenari2013recognizing})}\\
\hline
Multi-Part Bag-of-Poses~\cite{seidenari2013recognizing} & 82.00\\
Body part~\cite{devanne20143} & 87.04 \\
\hline
DAE&82.51\\
DAE\_CC&86.42\\
DAE\_TC&85.93\\
\textbf{DAE\_CTC} & \textbf{88.20}\\
\hline
\end{tabular}
}}%
}%
\end{table}%
\subsubsection{Comparison with DTW}%
\begin{figure}[tb]%
\centering%
\includegraphics[width=0.977\linewidth]{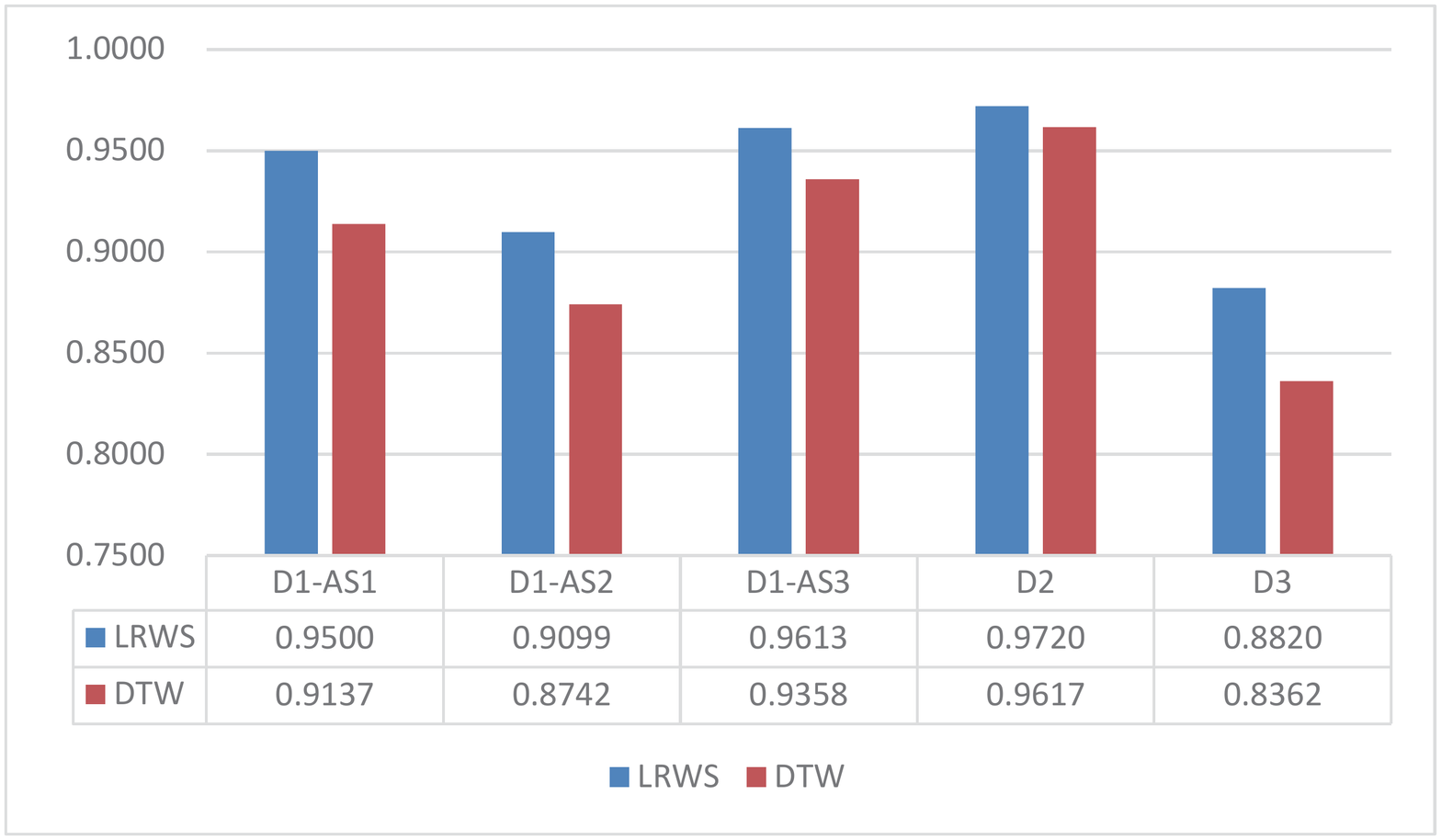}
\caption{Average recognition rates on MSR-Action3D, UTKinect-Action and Florence3D-Action using LRWS and DTW.}
\label{fig:LRWS}%
\end{figure}

We also make a comparison of the experimental results between the LRWS and DTW.
Instead of LRWS, we apply DTW to compute the phantom action templates to performe temporal registration.
Figure~\ref{fig:LRWS} reports the average recognition rates on \mbox{MSR-Action3D,} \mbox{UTKinect-Action,} \mbox{and Florence3D-Action} using LRWS and DTW.
In order to succinctly present this illustration, we number the three data sets \emph{D1}, \emph{D2}, and \emph{D3}, respectively.
LRWS on average performs 3\% better than DTW.%
\begin{figure}[tb]%
\centering%
\includegraphics[width=0.977\linewidth]{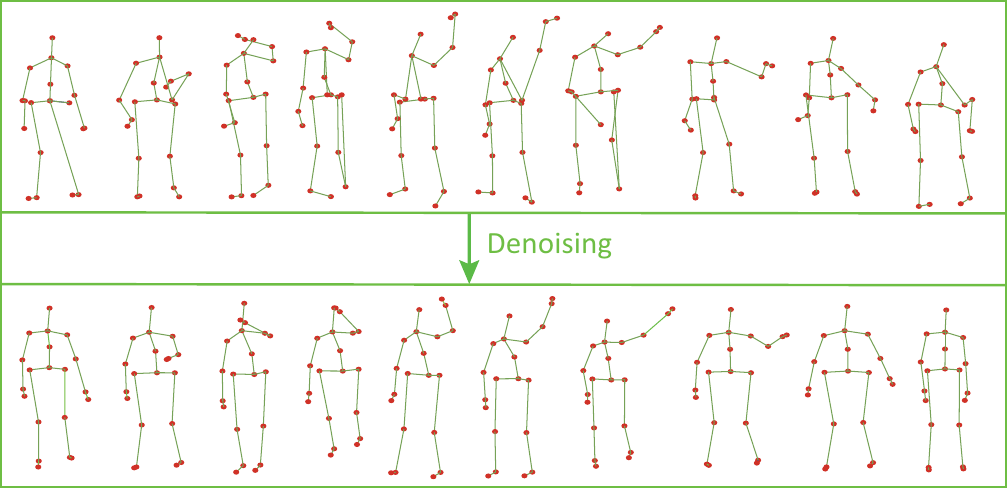}%
\caption{Examples showing the capability of our model to denoise corrupted data.}%
\label{fig:denoise}%
\end{figure}%
\subsubsection{Capability to restore corrupted skeleton}%
The DAE\_CCT not only has a strong ability of learning skeleton representation, but also has capability to restore realistic data from corrupted input.
In this section, we select an action instance and illustrate the restore result.
We first corrupt the instance by adding Gaussian noise and leaving out joints randomly.
Then the DAE\_CCT is used to reconstruct a new version.
In Figure~\ref{fig:denoise}, the top row is a skeleton sequence of \emph{``high arm wave''}, and the bottom row is the reconstruction one.
Compared with the above skeleton sequence, we can find that the noisy and the lost joints are well restored in the bottom one and the motions are more natural and fluent.%

\section{Conclusion and Future Work}%
\label{sec:7}
In this paper, a new framework for human action recognition is presented.
We propose a stacked denoising autoencoder to enhance the traditional deep neutral networks to learn and reconstruct the representation of human skeletons.
By integrating privileged information of action category and temporal coordinate into the learning architecture, the acquired features are more discriminative, able to capture the subtle but significant differences between actions.
We further propose a new temporal registration called LRWS to mitigate the nuisance variation of temporal misalignments.
We combine this with an approach based on the Fourier temporal pyramid (FTP) to perform temporal modeling and classification on the registered feature sequences.
We experimentally show that our method performs better than several commonly-used skeletal representations and it is comparable to state-of-the-art skeleton-based human action recognition approaches.
In addition, it is able to restore proper action sequences from highly noisy input data.

In our future work, we will explore an effective end-to-end solution for modelling temporal dynamics from the perspective of the whole action sequences.
In addition, we will consider other kinds of data, such as \mbox{RGB images} and depth maps, to enrich the representation of action sequences.%
\begin{sloppypar}%
\noindent\textbf{Acknowledgements.}~
We acknowledge support from %
the Youth Project of the Provincial Natural Science Foundation of Anhui Province %
1908085QF285 and 
1908085MF184, 
the Scientific Research and Development Fund of Hefei University %
19ZR05ZDA, 
the Talent Research Fund Project of Hefei University %
18-19RC26, 
the Key Research Project of Anhui Province %
201904d07020002, 
the National Natural Science Foundation of China under grants~61673359.
\end{sloppypar}%
\begin{sloppypar}%
\noindent\textbf{Disclosure.}~
All authors declare that there are no conflicts of interests. %
All authors declare that they have no significant competing financial, professional or personal interests that might have influenced the performance or presentation of the work described in this manuscript. %
Declarations of interest: none. %
All authors approve the final article.%
\end{sloppypar}%
\section*{References}%
\bibliography{reference}%
\end{document}